%% file: main.tex
\documentclass{article}

\PassOptionsToPackage{numbers, compress}{natbib}

\usepackage[preprint]{neurips_2026}

\makeatletter
\renewcommand{\@noticestring}{}
\renewcommand{\@bottomtitlebar}{
  \vskip 0.25in
  \vskip -\parskip
  \hrule height 1\p@
  \vskip -0.12in%
}
\makeatother

\input{preamble}

\title{Remember to be Curious: Episodic Context and Persistent Worlds for 3D Exploration}

\author{%
  \textbf{Lily Goli}$^{1,3,4}$ \quad
  \textbf{Justin Kerr}$^{2}$ \quad
  \textbf{Daniele Reda}$^{3}$ \quad
  \textbf{Alec Jacobson}$^{1,4}$ \\
  \textbf{Andrea Tagliasacchi}$^{1,3,5}$ \quad
  \textbf{Angjoo Kanazawa}$^{2}$ \\
  \\
  $^{1}$University of Toronto \quad
  $^{2}$UC Berkeley \quad
  $^{3}$Wayve \quad
  $^{4}$Vector Institute \quad
  $^{5}$Simon Fraser University \\
  \texttt{lily.goli@mail.utoronto.ca} \\[4pt]
 \href{https://recuriosity.github.io/}{{\color{recuriositypink}\bfseries recuriosity.github.io}}
}

\begin{document}

\maketitle
\input{fig/teaser}
\input{sec/0_abstract}

\input{sec/1_intro}

\input{sec/4_method}

\input{sec/5_results}
\input{sec/2_related}

\input{sec/6_conclusion}
\input{sec/7_acks}

\bibliographystyle{unsrtnat}
\bibliography{main}
\clearpage
\appendix
\input{sec/X_supp}

\end{document}

%% file: preamble.tex
\usepackage[utf8]{inputenc} %
\usepackage[T1]{fontenc}    %
\usepackage[pagebackref, breaklinks]{hyperref} 
\usepackage{url}            %
\usepackage{booktabs}       %
\usepackage{amsfonts}       %
\usepackage{nicefrac}       %
\usepackage{microtype}      %
\usepackage{xcolor}         %
\usepackage{wrapfig}
\usepackage{caption}
\usepackage{overpic}
\usepackage{multirow}
\usepackage{comment}
\usepackage{makecell}
\usepackage{soul}
\usepackage{microtype}
\usepackage{amsmath}
\usepackage[inline]{enumitem}
\usepackage{lipsum}

\usepackage{soul}
\usepackage{nicefrac}
\setuldepth{foobar}
\definecolor{recuriositypink}{RGB}{179,62,116}

\usepackage[capitalize]{cleveref}
\Crefname{section}{Section}{Sections}
\crefname{section}{Sec.}{Secs.}
\Crefname{appendix}{Appendix}{Appendices}
\crefname{appendix}{App.}{Apps.}
\Crefname{table}{Table}{Tables}
\crefname{table}{Tab.}{Tabs.}
\Crefname{figure}{Figure}{Figures}
\crefname{figure}{Fig.}{Figs.}
\Crefname{equation}{Eq.}{Eqs.}
\crefname{equation}{}{}

\definecolor{seen}{HTML}{858EF6}
\definecolor{unseen}{HTML}{D7A60F}

\newcommand{\img}{I}
\newcommand{\pred}{\hat{I}}
\newcommand{\action}{a}
\newcommand{\actions}{\mathcal{A}}
\newcommand{\obs}{o}
\newcommand{\obsall}{\tilde{o}}
\newcommand{\state}{s}
\newcommand{\pose}{p}
\newcommand{\depth}{D}

\newcommand{\predobs}{\hat{o}}

\newcommand{\policy}{\pi}
\newcommand{\env}{\mathcal{E}}
\newcommand{\fmodel}{\mathcal{F}}

\newcommand{\creward}{r^{\mathrm{cur}}}
\newcommand{\distance}{d}

\newcommand{\gs}{\mathcal{G}}
\newcommand{\gaussian}{g}
\newcommand{\gaussianproj}{\gaussian^{2D}}

\newcommand{\opacity}{\alpha}

\newcommand{\render}{\mathcal{R}}

\newcommand{\x}{\mathbf{x}}
\newcommand{\pixel}{\mathbf{u}}
\newcommand{\colr}{\mathbf{c}}
\newcommand{\covmat}{\Sigma}
\newcommand{\transmittance}{T}
\newcommand{\numgaussians}{L}

\newcommand{\prederr}{e}
\newcommand{\blur}{\mathcal{B}}

\newcommand{\downsamples}{\mathcal{D}_{s}}
\newcommand{\pixelset}{\Omega_s}
\newcommand{\rewardnew}{r_{\mathrm{new}}}
\newcommand{\rewardold}{r_{\mathrm{old}}}
\newcommand{\rewardthresh}{\tau}

\newcommand{\token}{z}
\newcommand{\temptoken}{\bar{z}}
\newcommand{\globaltoken}{\tilde{z}}
\newcommand{\crossatten}{\mathrm{CrossAttn}}
\newcommand{\selfatten}{\mathrm{SelfAttn}}
\newcommand{\linearatten}{\mathrm{LinAttn}}
\newcommand{\update}{\mathrm{Update}}
\newcommand{\qtoken}{q}
\newcommand{\patch}{u}
\newcommand{\dino}{f}
\newcommand{\memstate}{h}
\newcommand{\attnwindow}{W}

\newcommand{\policyparams}{\theta}
\newcommand{\valuefn}{V}
\newcommand{\behavior}{\mu}
\newcommand{\uniform}{\mathcal{U}}
\newcommand{\oldpolicy}{\policy_{\policyparams_{\mathrm{old}}}}
\newcommand{\advantage}{A}
\newcommand{\ret}{R}
\newcommand{\pporatio}{\rho}
\newcommand{\clipparam}{\epsilon}
\newcommand{\entropy}{\mathcal{H}}
\newcommand{\ppoloss}{\mathcal{L}_{\mathrm{PPO}}}
\newcommand{\criticloss}{\mathcal{L}_{\mathrm{critic}}}
\newcommand{\entropyloss}{\mathcal{L}_{\mathrm{entropy}}}
\newcommand{\expect}{\mathbb{E}}
\newcommand{\mixcoef}{\beta}

\renewcommand{\paragraph}[1]{\vspace{.25em}\noindent\textbf{#1.}}

%% file: fig/teaser.tex
\begin{figure}[ht!]
\begin{center}
\vspace{-3em}
\centering 
\includegraphics[width=\textwidth]{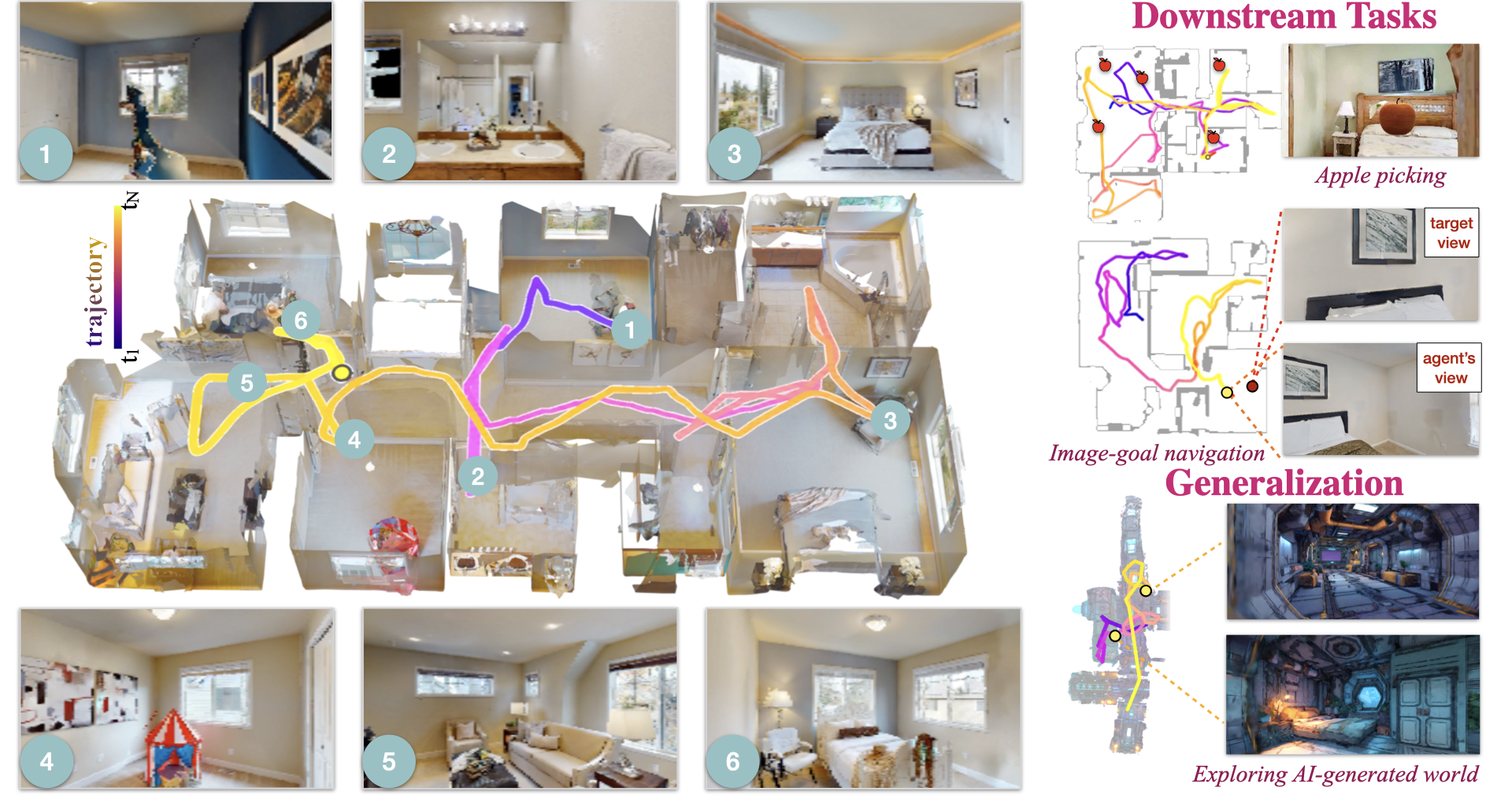}
\vspace{-1.5em}
\caption{
\textbf{Remember to be Curious.}
{Test-time trajectories of our curiosity-driven policy trained with episodic context and a persistent world model. \textbf{Left:} Our agent explores an indoor scene purely from an image stream; waypoints show the agent's views along the path. %
 \textbf{Right:} Our end-to-end design enables fine-tuning to downstream sparse-reward tasks %
 and zero-shot generalization to AI-generated worlds. Note: the model requires no explicit mapping at test time; trajectories are overlaid on bird's-eye-view for visualization  only. %
}
}
\label{fig:teaser}
\end{center}
\vspace{-1em}
\end{figure}

%% file: sec/0_abstract.tex
\begin{abstract}
Exploration is a prerequisite for learning useful behaviors in sparse-reward, long-horizon tasks, particularly {within 3D environments}. Curiosity-driven reinforcement learning addresses this via intrinsic rewards derived from the mismatch between the agent's predictive model of the world and reality. However, translating this intrinsic motivation to complex, photorealistic environments  remains difficult, as agents can become trapped in local loops and receive fresh rewards for revisiting forgotten states. In this work, we demonstrate that this failure stems from a lack of spatial persistence and episodic context. We show that effective curiosity requires a model of the world that is persistent and continuously updated, paired with an agent that maintains an episodic trajectory history to navigate toward novel regions. We achieve this using an online 3D reconstruction as a persistent model of the world, while the agent policy is parameterized as a sequence model over RGB observations to maintain episodic context. This design enables effective exploration during training while allowing the agent to navigate using solely RGB frames at deployment. Trained purely via curiosity on HM3D, our agent outperforms active-mapping baselines and generalizes zero-shot to Gibson and AI-generated worlds. Our end-to-end policy enables efficient adaptation to downstream tasks, such as apple picking and image-goal navigation, outperforming from-scratch baselines.

\end{abstract}

%% file: sec/1_intro.tex
\section{Introduction}
\label{sec:intro}
\vspace{-1em}
In his seminal work on latent learning, Edward Tolman demonstrated that agents---such as rats navigating a maze---can acquire complex knowledge of their environment even in the absence of explicit rewards~\cite{tolman1948}. 
This suggests that intelligent agents possess an inherent drive to explore, with curiosity serving as a mechanism for engaging with novel and uncertain environments. This phenomenon is intuitively observable in any toddler at a playground: they are never paralyzed by the absence of an explicit goal. Rather than waiting for a sparse external reward, it is far more natural for them to engage in unstructured play, freely wandering and exploring their surroundings.

In the context of artificial agents, this exploratory drive has the practical benefit of densifying sparse rewards in long-horizon tasks, such as visuomotor navigation. Despite the progress in goal-conditioned navigation~\cite{zhu2017, ddppo}, in real-world scenarios, these tasks fundamentally carry \textit{sparse} success signals.
Curiosity-driven exploration offers a compelling alternative, in which the agent derives intrinsic reward from surprise, a signal produced by the prediction error of a world model --- trained alongside the agent --- to anticipate the consequences of its actions~\cite{schmidhuber1991curiosityBoredom, pathak2017curiosity}.
However, scaling curiosity with end-to-end policies to complex photorealistic environments remains challenging, as agents frequently collapse into repetitive behavior driven by cyclic curiosity rewards.

In this work, we show that enabling curiosity requires both a persistent model of the world and an agent equipped with episodic context. This dual requirement addresses a two-fold problem of ``amnesiac'' exploration:
without historical context, agents repeatedly revisit the same locations; simultaneously, without a persistent and continuously updating world model, predictive errors spuriously arise in these revisited areas, yielding false novelty rewards for forgotten states.

\begin{wrapfigure}[10]{r}{1.2in}
    \vspace{-9pt}
    \centering \includegraphics[width=1.\linewidth]{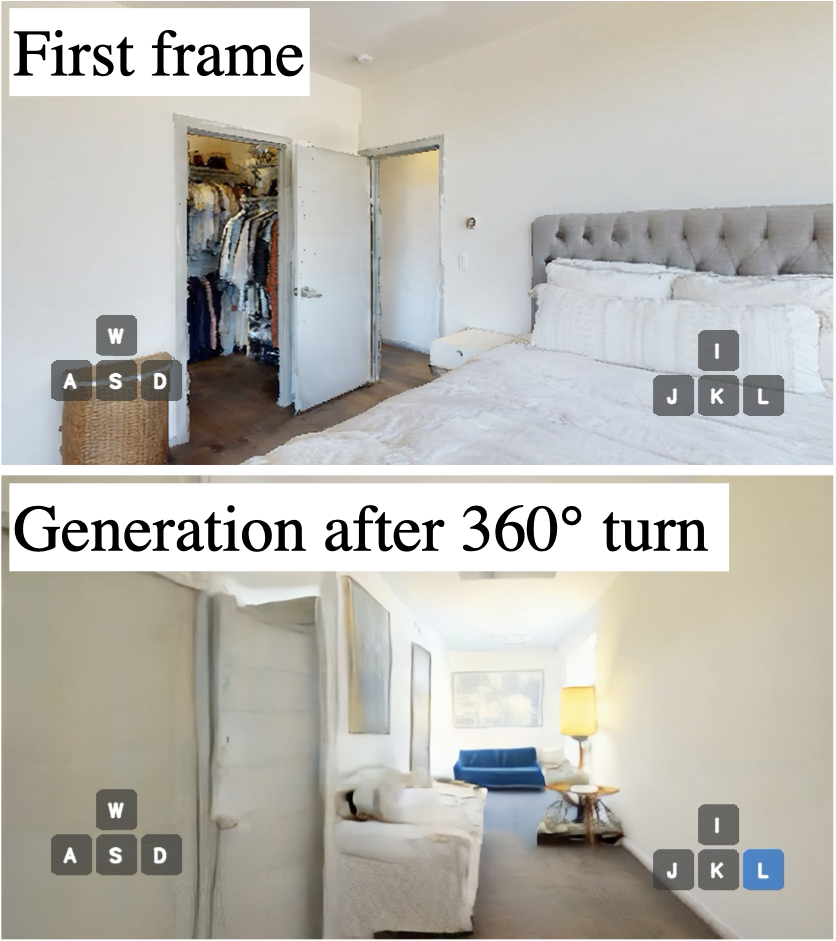}
    \vspace{-6pt}
\end{wrapfigure}
To study this, we focus on agents exploring static 3D photorealistic scenes. Ideally, the world model driving curiosity should be dynamic and continuously refined through lifelong agentic experience---a paradigm where action-conditioned video models show promise. However, spatial persistence remains a critical bottleneck for these models; for instance, LingBot-World~\cite{lingbotworld2026} fails to maintain scene consistency, generating an entirely different image after a simple in-place 360$^\circ$ turn (see inset figure). 
We therefore utilize a state-of-the-art online 3D reconstruction method (3DGS) as a proxy, as it enables spatial persistence while continuously updating itself online.
Traditional methods like ICM~\cite{pathak2017curiosity} lack this property, as their learned world model acts as a statistical prior over lifelong experience rather than an episodic record of the environment. Indeed, our experiments confirm that when the persistent memory is artificially capped to a short-term window, exploration capability deteriorates significantly.

On the agent side, we build a transformer-based policy that operates {purely} on a sequence of RGB observations. {By conditioning the policy on the agent's recent RGB history, we enable it to form its own internal representation of the episodic context directly from pixels.} %
This is in direct contrast to existing methods that bypass the exploration challenge by embedding \textit{explicit geometric maps} to guide the agent~\cite{ans2020,occant2020,gleam2025}, sometimes relying on ground-truth depth observations. While explicit mapping guarantees spatial awareness, such modularity often abstracts away rich semantic information, sacrifices the flexibility of end-to-end learning, and limits generalization to geometrically definable tasks.
Our image-sequence agent, conversely, requires no depth sensor or localization at deployment and adapts flexibly to semantically diverse downstream tasks.
Furthermore, we show that scheduled, temporary injections of random behavior into RL training suffice to overcome the reward-less stretches of long-horizon exploration~\cite{ecoffet2021goexplore}.
This enables an emergent behavior where the agent learns to traverse already-seen regions to find novel ``branches'' of unseen places.
Crucially, this is achieved without the imitation learning bootstrapping~\cite{chen2019learning}, or hierarchical goal selection of prior work~\cite{ans2020,occant2020,gleam2025}, highlighting curiosity's potential in sparse-reward settings.

Trained purely from curiosity on Habitat Matterport 3D~(HM3D)~\cite{ramakrishnan2021hm3d}, our agent outperforms RL-based active mapping baselines and generalizes zero-shot to Gibson~\cite{xiazamirhe2018gibsonenv}, and to out-of-distribution scenes from an entirely AI-generated 3D world~\cite{worldlabs} never encountered during training. Crucially, because our policy does not require explicit mapping at test time, it remains fully end-to-end and highly adaptable. Fine-tuned for just a few episodes on external task rewards, it outperforms policies trained from scratch. 
Our work demonstrates that coupling a persistent model of the world with an episodic agent provides practical insights into how curiosity-driven exploration can be scaled to complex, photorealistic 3D environments.

%% file: sec/4_method.tex
\section{Method}
\label{sec:method}
Assuming a static 3D scene, we aim to train a camera-agent to efficiently explore its surroundings purely from visual inputs. We wish for novelty-seeking behavior to arise without task-specific rewards or external exploration labels, and
so we formulate exploration as a self-supervised RL problem with a curiosity-driven learning signal derived from the agent's own experience. We defer discussion of \textbf{related works} and \textbf{background} to \Cref{sec:related} and Appendix~\ref{sec:background}, and describe our method first.

Our method has two key parts: a long-context transformer-based agent architecture which enables long-horizon exploration behavior, and a curiosity module based on 3D reconstruction which provides a stable reward for the agent to optimize with on-policy RL. The agent (\cref{fig:method}, left) takes in \textit{only} a stream of RGB observations from the 3D environment at hand, and outputs local camera motion per-timestep. It maintains internal memory long enough to enable it to discover exploration strategies like backtracking. 
The curiosity module (\cref{fig:method}, right) builds a photorealistic 3D scene reconstruction from the agent's stream of observations, then uses the disagreement between novel view renders and ground truth observation as reward. To optimize this curiosity-based reward the agent necessarily must learn to visually explore 3D scenes and navigate complex environments to seek novel viewpoints.

\paragraph{Problem setup} We frame exploration as interaction with an environment.
At each time $1{<}t{<}N$, the agent takes action $\action_t$ to move through environment $\env$ and receives new observation $\obs_{t+1}$, transitioning from its current state to the next.
While observations can include privileged sensory information at training time (denoted $\obsall_{t+1}$ when so augmented), the agent relies solely on a visual RGB stream at test time, enabling easy and general deployment. 
{Specifically, the privileged inputs required at training time are the camera pose and depth image.}

\input{fig/method}
\subsection{Persistent 3D forward model}
\label{sec:forward-model}

For the agent to learn exploratory behavior, it should be rewarded for seeking novel regions of the environment. We achieve this via intrinsic curiosity by introducing an auxiliary model of the world during training.
This \textit{forward model} is tasked with predicting the next observation conditioned on an action, given the stream of observations so far:
\begin{equation}
    \predobs_{t+1} = \fmodel(\action_t, \obsall_{1:t}).
\end{equation}
The forward model $\fmodel$ defines a curiosity reward that activates upon encountering novel regions not yet explained by this model, by comparing its prediction to the observed reality:
\begin{equation}
    \creward_{t} = \distance(\predobs_{t+1}, \obs_{t+1}),
    \label{eq:curiosity-reward-distance}
\end{equation}
where $\distance$ is a visual prediction-error metric. A larger discrepancy indicates the new observation is poorly explained by the current forward model, corresponding to higher novelty.

We instantiate the forward model as an online 3D Gaussian Splatting (3DGS) model of the world, capable of incorporating streaming observations while maintaining a persistent model of what has been seen. We later show in \Cref{sec:experiments} that this persistence is crucial for a reliable reward signal. We refer the reader to Appendix~\ref{sec:background-3dgs} for background on 3DGS.
Given the training-time privileged stream of observations including RGB image, depth image and camera poses (i.e., $\obsall_{1:t} = \{(\img_i,\depth_i, \pose_i)\}_{i=0}^{t}$), we maintain a 3DGS representation $\gs_t$. 
For each observed RGB-D frame, a new gaussian per pixel is added to the scene, initialized from its color, depth, and camera pose. At fixed intervals, the representation is optimized on randomly selected past frames to improve reconstruction quality, and pruned and densified following~\cite{mcmc}. %

At each step, after the agent acts and the camera moves following environment dynamics, the forward model is queried at the new pose $\pose_{t+1}$ to render a prediction of the incoming observation: 

\begin{equation} \pred_{t+1} = \fmodel(\pose_{t+1} | \img_{1:t}, \depth_{1:t}, \pose_{1:t}) = \render(\gs_t, \pose_{t+1}), \end{equation}

where $\render$ denotes the differentiable 3DGS renderer. Since $\gs_t$ incorporates only observations up to time $t$, the prediction error primarily reflects whether the incoming view contains information not yet captured by the forward model.
To avoid rewarding reconstruction error near high-frequency details in the scene that are not meaningful for exploration, we compute prediction error after low-pass filtering and down-sampling both the prediction and observation:
\begin{equation}
    \prederr_t
    =
    \frac{1}{|\pixelset|}
    \sum_{(j,k)\in\pixelset}
    \left\|
    \downsamples(\blur(\img_{t+1}))^{j,k}
    -
    \downsamples(\blur(\pred_{t+1}))^{j,k}
    \right\|_2^2,
\end{equation}
where $\blur$ is a low-pass filter, $\downsamples$ is down-sampling operator by a factor of $s$, and $\pixelset$ is the down-sampled pixel set.
In our implementation, the distance $\distance$ in \cref{eq:curiosity-reward-distance} is defined by thresholding the above filtered error to provide a binary reward: %
\begin{equation}
    \creward_t
    =
    \begin{cases}
    \rewardnew, & \prederr_t > \rewardthresh, \\
    \rewardold, & \prederr_t \leq \rewardthresh,
    \end{cases}
\end{equation}
where $\rewardnew > 0$ rewards sufficiently unexplained views, and $\rewardold < 0$ gives a small penalty for views already explained by the forward model.

\subsection{Agent Architecture} 
\label{sec:agent}

Our agent, decoupled from the forward model, is tasked with seeking novelty and therefore must maintain its own record of past observations to take actions that lead to unseen areas. While a geometric map is an effective abstraction for this, it discards all information beyond scene geometry. {Instead we let the agent learn its own internal representation of the world,} %
where both geometry and semantics are preserved. To this end, we use a transformer backbone whose actions are conditioned on the full sequence of past visual observations and actions throughout the exploration episode, hence giving the agent the \textit{episodic memory} needed to learn a novelty-seeking policy $\policy$:
\begin{equation}
    \action_t \sim \policy(\cdot \mid \obs_{1:t}, \action_{1:t-1}).
\end{equation}
This memory has the further benefit of freeing the agent from being conditioned on any explicit geometric representation that abstracts the observations, making it straightforwardly adaptable to downstream tasks of varied nature that can benefit from exploratory behavior, through fine-tuning.

Concretely, the agent's episodic memory is represented as the sequence of past RGB observations and actions. This sequence is fed into a transformer backbone connected to the actor and critic heads that output an action distribution and a value estimate for the state at each timestep. The architecture, shown in~\cref{fig:method}, consists of an image encoder, causal attention layers over a temporal sliding window, and a global linear-attention memory module.
At time $t$, the input is $\{(\img_i, \action_{i-1})\}_{i=1}^{t}$, where each observation $\img_i$ is paired with the preceding action $\action_{i-1}$.
Each action is encoded geometrically as a Pl\"{u}cker-ray image~\cite{lfn} representing the intended camera transformation, independent of actual environment dynamics, and concatenated channel-wise with its corresponding RGB observation to form a unified RGB-action input.

This RBG-action image in each timestep is first compressed into a frame token through an image encoder. We also take the RGB image processed by DINOv2~\cite{dino} to provide richer visual features. 
A learnable query token cross-attends to the patch tokens and DINOv2 features:
\begin{equation}
    \token_i = \crossatten(\qtoken, [\patch_i, \dino_i]),
\end{equation}
where $\patch_i$ denotes RGB-action patch tokens, $\dino_i$ denotes DINOv2 features, and $\qtoken$ is the learnable frame query.
The resulting frame tokens are processed by sliding-window causal temporal attention:
\begin{equation}
    \temptoken_{1:t} = \selfatten_{\attnwindow}(\token_{1:t}).
\end{equation}
Causality ensures that each token uses only past and current observations, while the window size $\attnwindow$ keeps computation scalable over long episodes.
Sliding-window attention provides efficient direct local context, but long-range information can only propagate indirectly through deeper layers. 
We therefore interleave selected temporal layers with a global linear-attention memory module inspired by TTT and LoGeR-style long-context architectures~\cite{sun2024learning,zhang2025loger}. 
The module reads from a running memory state before updating it with the current token:
\begin{equation}
    \globaltoken_i = \linearatten(\temptoken_i, \memstate_{i-1}),
    \qquad
    \memstate_i = \update(\memstate_{i-1}, \temptoken_i).
\end{equation}
The memory readout is added back into the temporal stream, giving the policy access to information beyond the sliding attention window without full attention over the episode.
Finally, the current token is passed through the actor and critic heads to output action probabilities $\policy_{\policyparams}(\action_t \mid \obs_{1:t}, \action_{1:t-1})$ and value estimate $\valuefn_{\policyparams}(\obs_{1:t}, \action_{1:t-1})$. Action is then sampled from the probability provided by the $\policy_{\policyparams}$ at both train and test times.

\subsection{Training and regularization}
\label{sec:ppo}

We optimize our actor-critic policy using PPO~\cite{ppo}. 
During training, rollouts are collected from the current policy, and the intrinsic reward $\creward_t$ defined in~\cref{sec:forward-model} is used as the only reward signal.
\begin{wrapfigure}[9]{r}{1.2in}
    \vspace{-8pt}
    \centering
    \includegraphics[width=1.\linewidth]{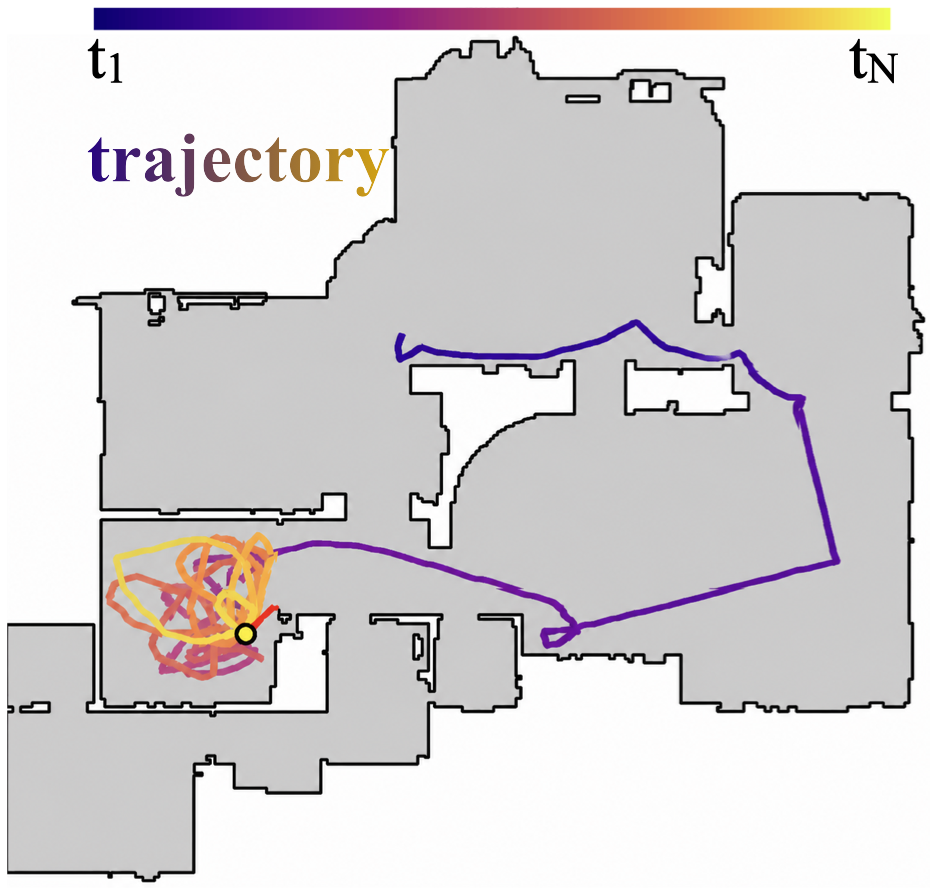}
    \vspace{-10pt}
\end{wrapfigure}
A practical difficulty in curiosity-driven exploration is maintaining action diversity after rewards become sparse. 
Once nearby regions have been explored, reaching new parts of the scene may require temporarily moving through already-seen areas without receiving any intermediate reward, such as backtracking to an unseen branch. 
Regularizing action entropy with an additional loss term alongside the PPO losses helps, but it is sensitive to its coefficient and the optimization landscape: too small a coefficient can be overwhelmed by other loss terms and allow exploration to collapse (see inset), while too large a coefficient can destabilize optimization. 
Inspired by SAPG~\cite{sapg2024}, which collects episode rollouts from a mixture of agents, we occasionally sample actions from a uniform random policy during the rollout collection in addition to a small entropy regularizer.

Specifically, at each training step, the executed action is sampled from a mixture of the learned policy and a uniform distribution over the discrete action set $\actions$:
\begin{equation}
    \behavior_{\policyparams}(\action_t \mid \obs_{1:t}, \action_{1:t-1})
    =
    (1-\mixcoef)\policy_{\policyparams}(\action_t \mid \obs_{1:t}, \action_{1:t-1})
    +
    \mixcoef \uniform(\action_t),
\end{equation}
where $\mixcoef$ is the probability of sampling from the uniform policy and $\uniform(\action_t)=1/|\actions|$. 
This guarantees persistent exploratory behavior during rollout collection without requiring the learned policy distribution itself to remain highly stochastic.
For the PPO update, we account for the mixed behavior distribution in the likelihood ratio by using the probability of the sampled action under the rollout behavior distribution in the denominator i.e. $\pporatio_t(\policyparams) = \nicefrac{\policy_\policyparams}{\behavior_{\policyparams_{\text{old}}}}$.
The resulting ratio is then used in the standard PPO clipped surrogate objective as described in Appendix~\ref{sec:background-ppo}, with the mixing coefficient annealed to zero over training (see Appendix~\ref{app:annealing}).

%% file: fig/method.tex
\begin{figure}
\centering
\includegraphics[width=\linewidth]{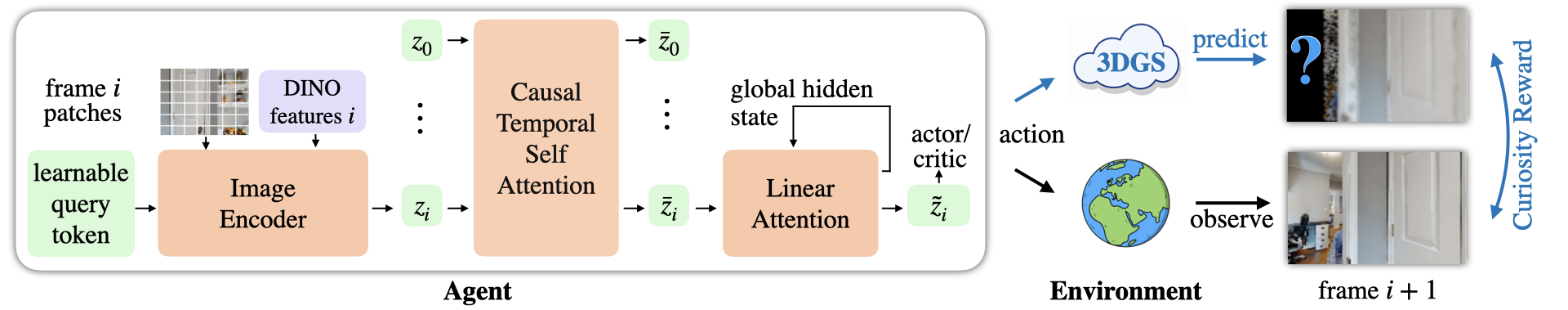}\\
\caption{\textbf{Method overview.}
{The \textbf{agent} (left) encodes each RGB frame into a per-frame token by fusing patch embeddings and DINOv2 features via a learnable query. Tokens are processed by causal temporal self-attention and a linear-attention module with a global hidden state, before an actor-critic head emits the next action and value estimate.
The \textbf{environment} (right) executes the action and returns the next observation; in parallel, a persistent 3DGS \emph{forward model} renders the same view from accumulated past experience, and the discrepancy with the observed view defines the \emph{curiosity reward}. The forward model is used only at training time (\textcolor[RGB]{78,159,246}{blue path}); at deployment the agent acts from RGB stream alone.}}
\label{fig:method}
\end{figure}

%% file: sec/5_results.tex
\section{Experiments}
\label{sec:experiments}
\label{sec:results}
In this section, we perform an experimental analysis of our curiosity-driven exploration. 
We first show how our agent can effectively explore indoor scenes achieving improved performance over previous RL-based active mapping methods~(\Cref{sec:coverage}).
To analyze the necessity of the agent's episodic memory and world persistence, we ablate memory capacity variations across both models in~(\Cref{sec:ablation}).
We then demonstrate how our agent, equipped with exploratory behavior, can readily be fine-tuned for downstream navigation tasks, such as apple-picking and image-goal navigation~(\Cref{sec:tasks}). 
Finally, we show our agent can generalize to AI-generated out-of-distribution scenes \textit{without} further training~(\Cref{sec:generalize}).

\input{fig/comparison}
\subsection{Indoor scene exploration --- {\Cref{fig:comparison}}}
\label{sec:coverage}
To assess the capability of our agent, we show test-time exploration of \textit{unseen} indoor scenes.

\paragraph{Implementation details}
We train our transformer backbone with Adam~\cite{adam} at a learning rate of $10^{-5}$ for $110$ million steps, with the random policy regularizer scheduled from $20\%$ to zero over $5$ million steps beginning at step $25$ million, and the action entropy coefficient decayed at a rate of $0.99$ from an initial value of $0.1$. Reward values are set to $r_{new}=0.5$, $r_{old}=2\times10^{-4}$.
The 3DGS primitives are initialized from depth and color at every step, optimized on $10$ random past views every $16$ steps with Adam~\cite{adam}, and densified via 3DGS-MCMC~\cite{mcmc}. Downsampling factor is set to $4$.
We simulate a drone agent with spherical embodiment spawned $1.25$ meters above the floor, and a $90^\circ$ FOV forward camera in Habitat~\cite{habitat}. It can perform four actions: \texttt{move forward} ($0.25$m), \texttt{look right} ($15^\circ$), \texttt{look left} ($15^\circ$), and \texttt{pause}.
Our drone agent is not constrained to the scene navmesh. Instead, it moves freely in 3D, and collisions are computed directly by ray-tracing against the scene mesh and checking contact with the agent’s spherical embodiment. This avoids the navmesh-dependent sliding shortcut learning identified by \citet{analyze}.
{Actions are executed deterministically unless the agent collides, in which case it halts in place.}
All train- and test-time episodes are run for 1024 steps. Training runs for $5.5$ days on a $8\times$ 80GB H100 GPUs.

\paragraph{Datasets}
We train on the HM3D~\cite{ramakrishnan2021hm3d} training set (consisting of $800$ scenes), with each episode initialized at a random location.
We evaluate on the HM3D validation set ($100$ scenes, $2$ random starting points each, $200$ episodes total) and, to assess generalization beyond HM3D, on Gibson~\cite{xiazamirhe2018gibsonenv} ($86$ mostly office-space scenes, $1$ starting point each).

\paragraph{Baselines}
We compare against RL-based exploration methods, as our goal is to evaluate learned policies rather than online planners optimizing per-scene \textit{geometric} objectives.
Specifically, we compare against \textbf{Active Neural SLAM (ANS)}~\cite{ans2020} and \textbf{Occupancy Anticipation (OccAnt)}~\cite{occant2020}, two representative map-driven RL approaches.
We exclude GLEAM~\cite{gleam2025}: despite being more recent, it follows the same map-based hierarchical paradigm and assumes panoramic observations, a different action space, and a test-time collision-unaware A* local planner, making it incompatible with our forward-facing RGB, collision-aware Habitat setting.
The \texttt{-RGB}, \texttt{-depth}, and \texttt{-RGBD} variants denote different input modalities to their respective RL policy module: ANS predicts a top-down map from RGB; ANS-depth projects depth directly; OccAnt-rgbd additionally predicts map completions beyond the projection; and OccAnt-rgb does so without depth. All baselines use default hyperparameters from the official OccAnt codebase and are trained for $110$M steps on the same dataset and simulator.

\paragraph{Metrics}
Following~\citet{activeneuralmap2023}, we measure exploration via 3D scene completeness at $3$ time horizons ($256$, $512$, and $1024$ steps): the percentage of points on the reachable ground truth mesh surface (sampled uniformly, $200$k points) whose closest observed point, back-projected using ground-truth depth, lies within $5$cm. We also report the average distance from each ground truth point to its nearest observed point.

\paragraph{Analysis}
Our agent achieves greater 3D completeness faster than all baselines, while requiring only RGB input at test time, unlike OccAnt-RGBD and ANS-depth, which require ground-truth depth and are thus more limited in deployment.
All baselines additionally rely on raw camera pose from agent sensors, subsequently corrected by a localization module, though this distinction is less critical in our {deterministic motion setting}.
As shown in \Cref{fig:comparison}, map-based methods exhibit characteristic failure modes: OccAnt and ANS-RGB can become locally trapped or stuck on the scene geometry due to erroneous map predictions, while ANS-depth over-invests in local coverage at the expense of long-horizon exploration efficiency.
Qualitatively, our agent exhibits emergent behaviors such as seeking doorways, traversing corridors, and returning to junctions, while almost never resorting to the \texttt{pause} action during exploration.

\input{fig/memory_ablations}
\subsection{Memory Ablations ---~\Cref{fig:memory_ablation}}
\label{sec:ablation}
We study the effect of memory in the forward model and in the agent. We start from the basic building blocks introduced by {ICM}~\cite{pathak2017curiosity} and gradually add to the memory capacities of each module.
We use the same HM3D evaluation dataset and metric setup as in~\Cref{sec:coverage}.

\input{fig/apples}
\paragraph{Baselines}
Starting from ICM~\cite{pathak2017curiosity}, we ablate the world model and agent memory independently. For the world model, we test ICM with its original inverse and forward dynamics modules, paired with either a capacity-matched RNN (\textbf{ICM-RNN}) as in~\cite{pathak2017curiosity}, or our transformer (\textbf{ICM-Transformer}), isolating the effect of the world model alone.
We then replace the forward model with a 3DGS limited to the last $64$ frames (\textbf{Short Memory 3DGS}) at each time step, ablating short-horizon world memory.
For agent memory, we compare a capacity-matched RNN, and transformers with limited context windows (ctx$=1, 4, 16$).
Finally, we ablate memory asymmetry between actor and critic (Actor w/ctx$=1$; Critic w/ ctx$=1$).

\paragraph{Analysis}
Persistent memory consistently benefits exploration across all modules. ICM without persistent world memory catastrophically \textit{collapses} spinning in place or getting stuck on geometry. {Adding a persistent 3DGS forward model to RNN makes significant gains, and non-persistent Short Memory 3DGS causes agents to roam locally without venturing far.}
Agent memory ablations confirm that both RNN and short-context transformers degrade exploration, and notably asymmetric memory, even with only the critic retaining context, outperforms having no memory at all.

\subsection{Task-based fine-tuning --- \Cref{fig:apples} and~\Cref{fig:image_goal} (left)}
\label{sec:tasks}
We show that our map-free agent architecture transfers readily to downstream navigation tasks via fine-tuning, without any architectural modifications, leveraging its exploratory pretraining to achieve stronger performance in sparse-reward settings.

We evaluate on two tasks.
\textbf{Apple-picking}~\cite{mirowski2016learning}: the agent must locate and approach apples scattered across the scene; picked apples are removed, and the agent receives a positive reward per pick and a small step penalty otherwise. Apples are placed at reachable positions sampled with probability proportional to their minimum distance to the already-selected set (i.e., avoiding clusters).
\textbf{Image-goal navigation}: the agent is given a target RGB image and must reach the corresponding viewpoint; success requires $\geq50\%$ of the target's 3D points to be visible and the agent to be within $1.5$m of the target pose.
The goal image is passed through image encoder and then provided as an additional token to the temporal module with dense attention to all other tokens. 

\paragraph{Experimental setting}
Both tasks use HM3D~\cite{ramakrishnan2021hm3d}. Apple-picking uses five apples per scene; image-goal navigation samples target poses reachable by the agent and sufficiently far from scene geometry.
We compare against a matched agent trained from scratch on the task reward only for $118$M steps; equal to our $110$M pretraining plus $8$M fine-tuning with external reward. We also report our non-fine-tuned exploration agent as a zero-shot baseline.
We report success rate: average percentage of apples picked and percentage of image goals reached across the test set.

\paragraph{Analysis}
Operating purely on RGB input without explicit geometric representations, our agent is flexible to adapt to both of the downstream semantic tasks through fine-tuning alone. 
In \textbf{apple-picking}, as shown in~\cref{fig:apples}, the baseline degrades under sparser reward (fewer apples at train-time), its search confined to one room, while our fine-tuned agent scales consistently with apple count.
Without fine-tuning, pure exploration hits some apples zero-shot. 
\textbf{Image-goal navigation} is a more extreme sparse-reward regime: the baseline agent collapses to a roaming-around behavior, while our fine-tuned model leverages its exploratory prior to find the target view more reliably;~\cref{fig:image_goal} (left).
This is due to the fact that broader exploration increases reward exposure during training, enabling the re-targeting of general exploration to goal-based navigation.

\input{fig/image_goal}
\subsection{Out-of-distribution scenes --- \Cref{fig:image_goal} (right)}
\label{sec:generalize}
To assess generalization beyond indoor Habitat scans, we apply our agent zero-shot to two AI-generated worlds from World Labs~\cite{worldlabs} (Hobbit World and Spaceship).
These scenes differ not only in aesthetic but also in rendering pipeline, as they are rendered from a 3DGS representation of these worlds rather than the realistic mesh scans of HM3D.
Nevertheless, the agent exhibits coherent exploratory behavior: it navigates corridors, discovers doors to new spaces, and avoids collisions.
Specifically, the agent encounters only $2$ and $3$ collisions  over $256$ steps in the two worlds, suggesting robust internal representations of {navigable} space.

%% file: fig/comparison.tex
\begin{figure*}[t]
\centering
\setlength\tabcolsep{0.01pt}
\includegraphics[width=\linewidth]{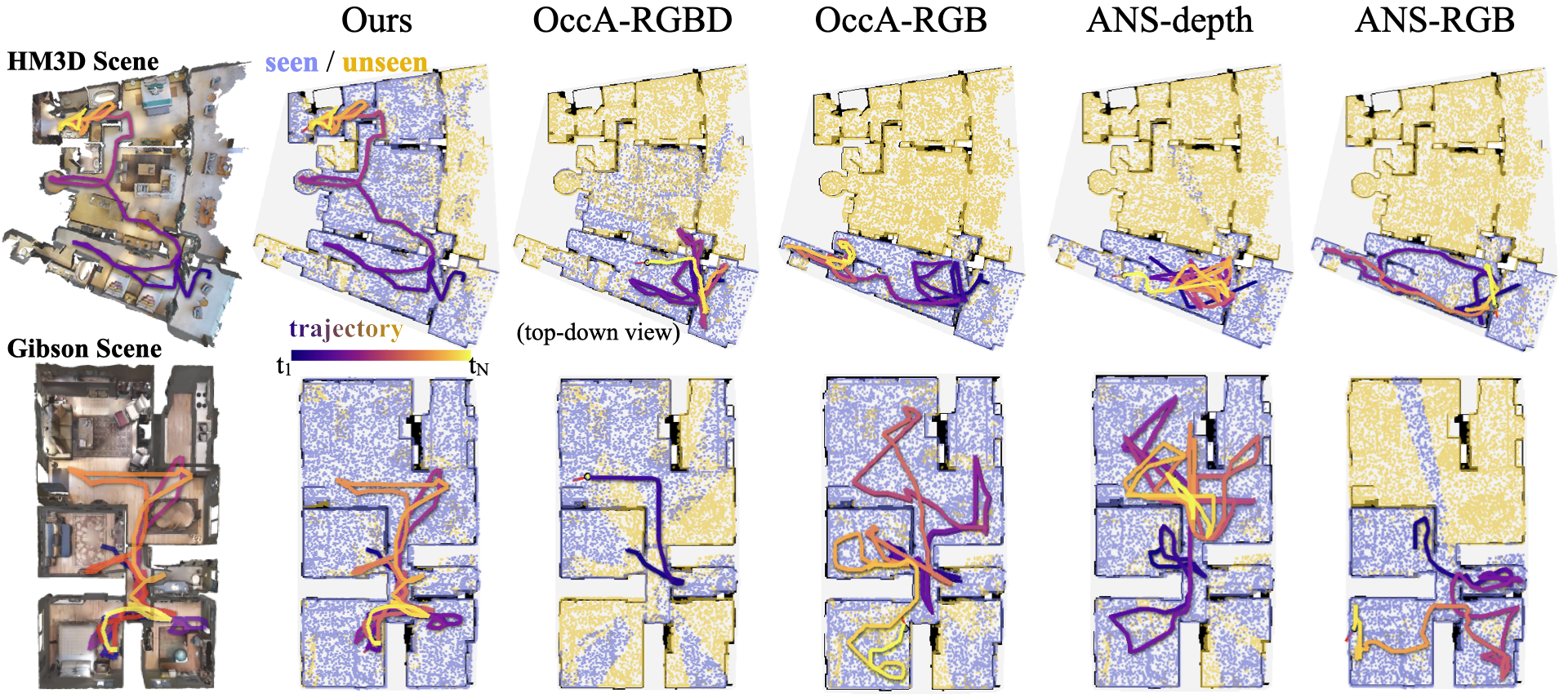}\\
\resizebox{\linewidth}{!}{
\setlength{\tabcolsep}{8pt}%
\begin{tabular}{lcccc|cccc}
\toprule
        & \multicolumn{4}{c}{\textbf{HM3D}} 
        & \multicolumn{4}{c}{\textbf{Gibson}} \\ \midrule
        & \multicolumn{3}{c}{Completeness \% $\uparrow$} 
        & Avg. dist. (m) $\downarrow$
        & \multicolumn{3}{c}{Completeness \% $\uparrow$} 
        & Avg. dist. (m) $\downarrow$ \\ 
Step    & @256 & @512 & @1024 & @1024
        & @256 & @512 & @1024 & @1024 \\ \midrule
ANS-RGB~\cite{ans2020}   & 45.28 & 54.68 & 65.39 & 0.41
          & 55.41 & 64.20 & 73.14 & 0.30 \\
ANS-depth~\cite{ans2020} & 51.02 & 61.45 & 69.68 & 0.34
          & 63.04 & 72.79 & 79.89 & 0.18 \\
OccA-RGB~\cite{occant2020}  & 47.67 & 58.32 & 68.86 & 0.33
          & 57.33 & 67.35 & 77.93 & 0.16 \\
OccA-RGBD~\cite{occant2020} & 52.71 & 64.91 & 74.62 & 0.18
          & 63.06 & 72.96 & 81.23 & 0.14 \\ 
\textbf{Ours}      & \textbf{56.5}  & \textbf{66.69} & \textbf{74.94} & \textbf{0.14}
          & \textbf{66.95} & \textbf{75.79} & \textbf{82.42} & \textbf{0.10} \\         
          \bottomrule
\end{tabular}%
}
\caption{
\label{fig:comparison}
Our agent explores the scenes more thoroughly leading to a higher completeness in 3D scene coverage in HM3D~\cite{ramakrishnan2021hm3d} and Gibson~\cite{xiazamirhe2018gibsonenv} datasets. Figure shows agents trajectories and top-down projection of 3D points as \textbf{\textcolor{seen}{seen}} or \textbf{\textcolor{unseen}{unseen}} by the agents.}
\end{figure*}

%% file: fig/memory_ablations.tex
\begin{figure*}[t]
\centering
\setlength\tabcolsep{0.01pt}
\includegraphics[width=\linewidth]{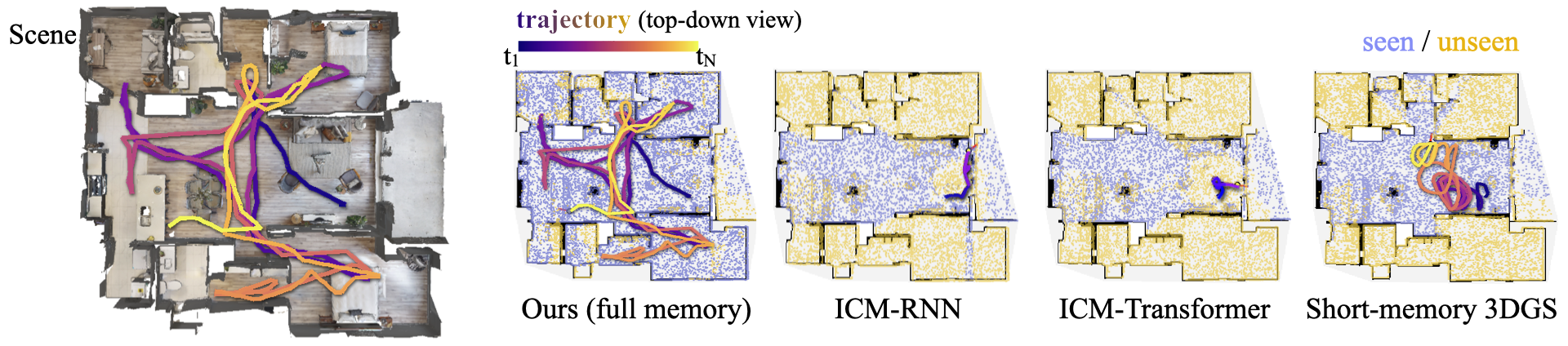}\\
\resizebox{\linewidth}{!}{
\setlength{\tabcolsep}{4pt}%

\begin{tabular}{lcccccccccc}
\toprule
        Metrics
        & \textbf{Ours}
        & \makecell{ICM \\ RNN}
        & \makecell{ICM \\ transformer}
        & \makecell{Short Memory \\ (64) 3DGS}
        & \makecell{Ours \\  RNN}
        & \makecell{Actor \\ w/ ctx=1}
        & \makecell{Critic \\ w/ ctx=1}
        & \makecell{Both \\ w/ ctx=16}
        & \makecell{Both \\ w/ ctx=4}
        & \makecell{Both \\ w/ ctx=1}
        \\ \midrule
 Compl. \%@256 $\uparrow$
        & \textbf{56.5}  & 32.70 & 35.25 & 33.35 & 50.70 & 51.95 & 47.09 & 51.80 & 45.87 & 40.66 \\
 Compl. \%@512 $\uparrow$
        & \textbf{66.69} & 35.52 & 38.89 & 36.05 & 59.52 & 61.79 & 56.09 & 60.66 & 53.83 & 46.09 \\
  Compl. \%@1024 $\uparrow$
        & \textbf{74.94} & 37.36 & 43.44 & 38.09 & 67.33 & 70.11 & 63.72 & 67.77 & 60.56 & 50.39 \\ \midrule
Avg. dist.  $\downarrow$
        & \textbf{0.14}  & 1.10  & 0.94  & 1.07  & 0.29  & 0.24  & 0.42  & 0.28  & 0.43  & 0.68  \\ \bottomrule
\end{tabular}%
}
\caption{\textbf{Ablations} on the memory capacity of the forward model and the policy network {on} HM3D~\cite{ramakrishnan2021hm3d}, show that a persistent forward model is necessary for effective exploration. Further, the agents equipped with memory capacity can achieve higher 3D scene coverage with less concentrated local loops in its trajectory.
}
\vspace*{-0.6cm}
\label{fig:memory_ablation}
\end{figure*}

%% file: fig/apples.tex
\begin{figure}
    \centering
\includegraphics[width=\linewidth]{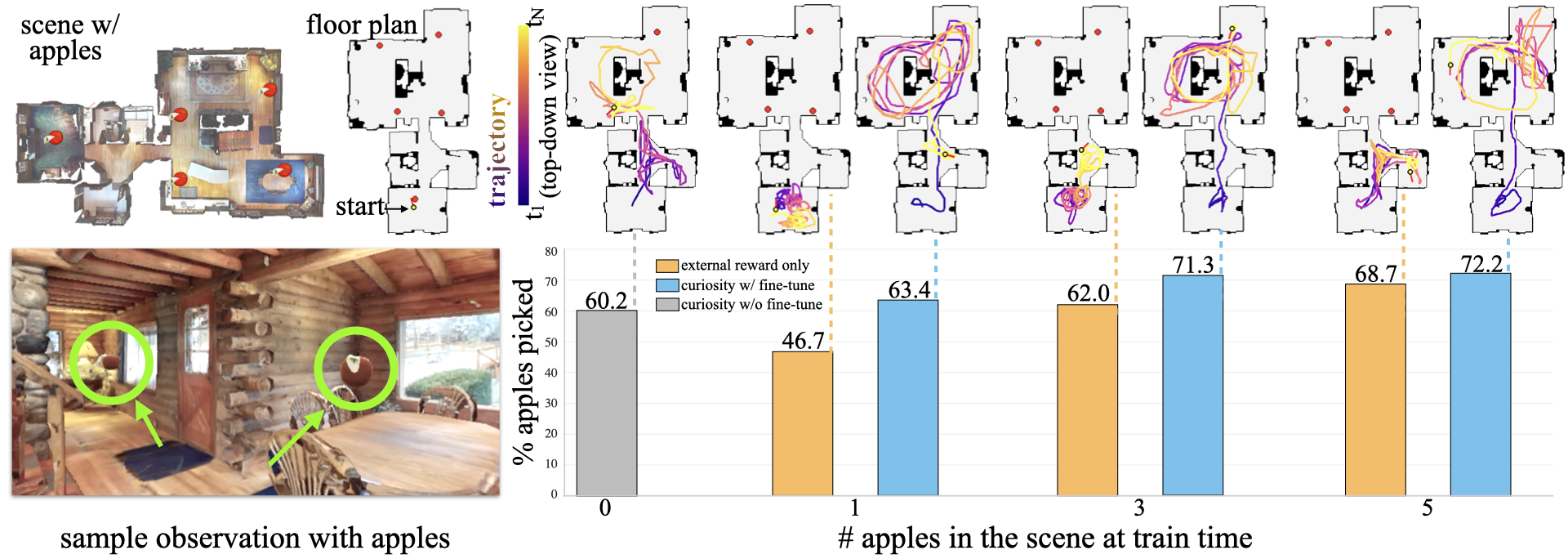}\\
\caption{After a few fine-tuning episodes on the apple-picking reward, our exploration agent achieves higher task success than an agent trained from scratch using only the external task reward with the same total number of episodes. This advantage is larger when fewer apples are available per training scene, where rewards are sparser, {and more reward-less exploration is needed.}%
}
\vspace*{-1em}
    \label{fig:apples}
\end{figure}

%% file: fig/image_goal.tex
\begin{figure}
    \centering
\includegraphics[width=\linewidth]{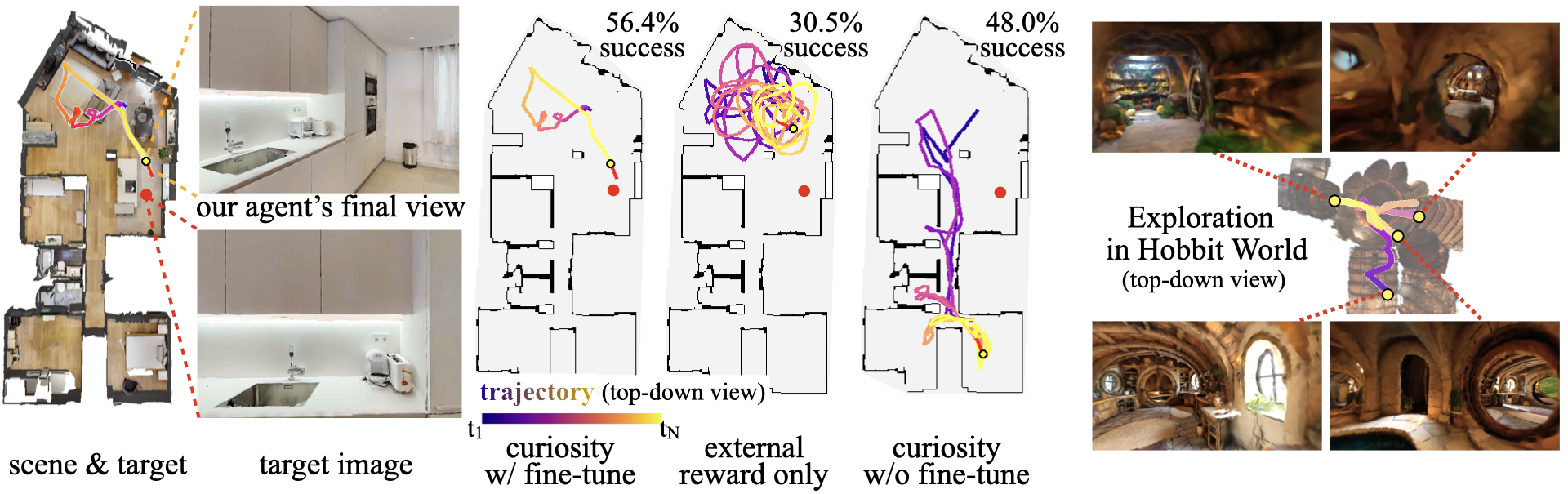}\\
\caption{\textbf{Left}: Our exploration agent, when fine-tuned for a few episodes on image-goal navigation reward, outperforms an agent trained from scratch on this reward alone, redirecting its general-purpose exploration toward a targeted objective. \textbf{Right}: Our agent generalizes zero-shot to AI-generated worlds of different aesthetic and  representation.}
    \label{fig:image_goal}
    \vspace*{-0.2cm}
\end{figure}

%% file: sec/2_related.tex
\section{Related Work}
\label{sec:related}
We survey the most related works here; a more comprehensive discussion is provided in Appendix~\ref{app:related_big}.

\paragraph{Intrinsic Motivation and Curiosity} 
Curiosity-driven exploration often rewards prediction errors in a learned forward dynamics model~\cite{pathak2017curiosity, burda2019rnd, pathak2019disagreement} or maximizes episodic visitation proxies~\cite{savinov2019episodic, badia2020ngu, henaff2022e3b, fu2023gobi}. However, these approaches struggle with non-stationary reward signals and fundamentally lack a persistent, improving model of the world~\cite{castanyer2024sofe}. Furthermore, standard exploration policies are typically reactive or limited to short recurrent contexts, preventing agents from planning based on accumulated history~\cite{ecoffet2021goexplore}. We address both limitations by grounding curiosity in a persistent 3D reconstruction paired with a full-history sequence model for long-horizon planning toward novelty.

\paragraph{Active Mapping and Next-Best-View} 
Traditional next-best-view (NBV) methods greedily select viewpoints to maximize geometric information gain~\cite{activeneuralmap2023, fisherrf2024, magician2026, llmfisher2025}. Similarly, RL-based mapping methods often condition policies on explicit geometric maps~\cite{ans2020, occant2020, gleam2025} or rely on human demonstrations~\cite{chen2019learning, deepexplorer2023}. By treating map-building as the end goal and explicitly conditioning on geometry, these approaches restrict their adaptability to semantically diverse tasks. Our policy, conversely, requires no explicit geometric map or depth sensor at deployment, operating purely from RGB streams to enable flexible downstream task adaptation.

\paragraph{3D Representations as World Models} 
Recent video-generative models offer a promising data-driven approach to modeling world dynamics~\cite{genie3, lingbotworld2026, ye2026dreamzero}, but currently suffer from severe spatial forgetting and open-loop generation~\cite{ma2026outofsight}. Alternatively, explicit 3D reconstructions like 3D Gaussian Splatting~\cite{kerbl2023gaussians, splatam} serve as rudimentary, dynamics-free world models that guarantee exact geometric consistency. Rather than claiming 3DGS uniquely solves this problem, we use it as a controlled proxy demonstrating that spatial persistence and closed-loop online updating are strictly required for reliable curiosity signals, with dynamic scenes as a natural next frontier as generative models mature.

%% file: sec/6_conclusion.tex
\section{Conclusions}
\label{sec:conclusions}
We introduced \textit{Remember to be Curious}, a framework for end-to-end curiosity-driven exploration of static realistic 3D scenes, where the agent operates solely from a visual input stream. %
Through using a persistent online 3D reconstruction model as a proxy for the forward model, we demonstrate the necessity of two components for long-horizon exploration:  a persistent forward model of the world for a reliable curiosity signal, and an episodic memory architecture for planning toward novelty.

While our use of 3DGS confines the agent to static scenes, we see promise in action-conditioned video models for addressing this limitation. We view our study as a guide for what these future world models require to enable intrinsic motivation. In particular, spatial persistence is key, alongside the ability to continuously update internal representations based on closed-loop online observations.
We hope this work opens avenues for curiosity-driven exploration across diverse real-world environments and embodiments, extending to dynamic settings as world modeling matures.

%% file: sec/7_acks.tex
\section{Acknowledgments}
\label{sec:acks}

We thank Tyler Bonnen, Antoine Guédon, Vincent Lepetit, Vassia Simaiaki, Jamie Shotton, and Alexei Efros for constructive feedback and helpful discussions. We also thank members of the DGP Lab at the University of Toronto, KAIR Lab at UC Berkeley, and Theia Lab at Simon Fraser University for their feedback, support, and discussions.

Our research is funded in part by NSERC Discovery (RGPIN–2022–04680), the Ontario Early Research Award program, the Canada Research Chairs Program, a Sloan Research Fellowship, the DSI Catalyst Grant program and gifts by Adobe Inc.

%% file: sec/X_supp.tex
\section{Video Results}
Please refer to our website, \href{https://recuriosity.github.io/}{{\color{recuriositypink}\bfseries recuriosity.github.io}}, for video results of our agent exploring diverse 3D environments and performing downstream tasks.
\input{sec/2_related_big}
\input{sec/3_background}

\section{Scheduled regularization annealing}
\label{app:annealing}
Since mixing the learned policy with a uniform random policy shifts the action distribution seen by the critic, it can bias its value estimates. Therefore, we gradually anneal
$\mixcoef$ from $0.2$ to zero over a scheduled window during training. 
This provides additional exploration when the policy is weak and curiosity rewards become sparse locally, while reducing the bias introduced by random actions as the learned policy improves. 
At test time, we set $\mixcoef=0$ and act directly from the learned policy.

%% file: sec/2_related_big.tex
\section{Related Work Discussion}
\label{app:related_big}

\paragraph{Reinforcement learning with intrinsic motivation}
In reinforcement learning (RL), agents learn by interacting with the environment and improving based on the rewards they receive. However, many real-world environments provide sparse or no extrinsic reward, motivating intrinsic-reward methods that encourage exploration through auxiliary objectives such as maximizing state coverage, information gain, or self-generated goals~\cite{seo2021re3, sukhija2024maxinforl, campero2021amigo}.

{Curiosity is a form of intrinsic reward, first framed by~\citet{schmidhuber1991curiosityBoredom} as reward for prediction error in a learned world model. ~\citet{pathak2017curiosity} proposed a deep RL curiosity framework in which a latent forward dynamics model is trained to predict consequences of actions and the agent is rewarded to seek states with large prediction error. Follow-up works stabilized this curiosity reward further with forward dynamics ensemble disagreement~\cite{pathak2019disagreement} or fixed random network representations~\cite{burda2019rnd}. However, these rewards lack episodic persistence: the forward dynamics model only updates through gradient steps across training, so the agent can repeatedly revisit surprising states within an episode to accumulate rewards without having them updated~\cite{badia2020ngu}, fixating on a narrow set of states while the forward model receives little diverse data to improve on. Further, the reward target becomes non-stationary as the learned latent representation of states evolve, making the signal noisy and optimization unstable~\cite{burda2019large, castanyer2024sofe}.}

To address this, follow-up works augment curiosity with episodic visitation counts~\cite{raileanu2020ride, badia2020ngu, fu2023gobi}, but count-based methods do not scale to continuous, high-dimensional observations~\cite{bigazzi2022focus, savinov2019episodic}, and efforts such as E3B~\cite{henaff2022e3b} that use compact learned embeddings inevitably abstract away geometric and semantic structure. 
Whether grounded in forward dynamics or visitation counts, no method provides a persistent, improving model of the world for a stable curiosity signal in an exploration episode.

Moreover, the exploration policy which is architecturally decoupled from the forward dynamics model~\cite{pathak2017curiosity} is either solely reactive to the current observation or limited to a short recurrent context, preventing the agent from planning exploration based on its accumulated history. 
{While our proposed 3D modeling of the world sacrifices a lifelong learned world model, we use it as a controlled proxy to study the role of persistence in curiosity-driven exploration.
}

\paragraph{Active mapping and next-best-view selection} 
{While curiosity-driven methods learn a policy that seeks novelty as a behavioral objective, active mapping and next-best-view (NBV) methods treat exploration as a means to an end: building a map of the scene. Classical and learned NBV approaches~\cite{activeneuralmap2023, fisherrf2024, goli2023, udp2022, macarons2023, nextbestpath2025, activesplat2025, activegamer2025, rtguide2024, mapex2025, pipeplanner2025} greedily select viewpoints by estimated information gain or coverage; MAGICIAN~\cite{magician2026} extends this to a more comprehensive tree search over information gain maximizing trajectories, and recent works combine LLMs with geometric planning for even longer-horizon goal selection and planning~\cite{llmfisher2025, wang2026explore}. These methods do not learn from experience, are scene-specific and replan from scratch per scene, making them difficult to extend to downstream tasks beyond mapping.}

{On the other hand, a line of RL-based methods learns a mapping policy conditioned on an explicit geometric map, giving the agent spatial but not semantic memory. \citet{chen2019learning} and~\cite{lodel2022look} learn exploration policies conditioned on egocentric depth-derived maps via imitation learning and coverage fine-tuning; ANS~\cite{ans2020} and OccAnt~\cite{occant2020} build hierarchical policies on SLAM-generated maps where a global RL module selects coverage-rewarded goals and a local planner executes them; and GLEAM~\cite{gleam2025} follows the same paradigm at larger scale with panoramic observations. By conditioning on a geometric map and optimizing for its coverage, these policies acquire no semantic understanding of the world and therefore can only be extended to tasks that can be defined geometrically, such as point-goal navigation. DeepExplorer~\cite{deepexplorer2023} forgoes the map via a short recurrent context but still relies on expert demonstrations, and GenNBV~\cite{chen2024gennbv} learns an object-centric NBV policy from occupancy and image features, targeting active object-level reconstruction rather than general exploration.}

{Our curiosity policy is conditioned on RGB observations alone, requiring no geometric map, depth sensor, or localization at test time, and is flexible to finetuning for downstream tasks such as image-goal navigation where exploratory behavior is a prerequisite.}

\paragraph{3D scene representations and world models}
{3D Gaussian Splatting~\cite{kerbl2023gaussians} reconstructs static scenes as explicit Gaussian primitives with fast differentiable rasterization, with follow-up work improving optimization speed and quality~\cite{mcmc, meuleman2025onthefly}. Online SLAM methods extend this to real-time incremental reconstruction from an image stream~\cite{splatam}. Feed-forward methods~\cite{wang2024dust3r, wang2025cut3r, zhang2025loger, wang2025vggt} enable fast dense reconstruction from sequential frames without per-scene optimization, though most output point clouds rather than renderable representations. These reconstruction methods can be viewed as a rudimentary form of a world model: given a camera action, they synthesize the resulting observation, approximating next-state prediction in a geometrically grounded but dynamics-free manner.}

{Video generative world models~\cite{genie3, lingbotworld2026, ye2026dreamzero} offer broader capacity by learning action-conditioned dynamics from data, showing promise for modeling non-stationary worlds for interactive tasks. However, current approaches suffer from spatial forgetting and temporal drift over long horizons~\cite{ma2026outofsight}, and approaches that address this with persistent 3D memory~\cite{lyra2026, zhou2025persistent3d, wu2025video} mostly revert to a static world assumption to maintain consistency. Further, most generative methods condition on previously generated video chunks rather than real observations, leaving planning open-loop.}

{We show that curiosity-based exploration requires a forward model grounded in real, closed-loop observations with persistent memory. Rather than claiming this is uniquely solved by 3DGS, we use it as a controlled instantiation for static indoor scenes to study what properties a forward dynamics model must have to support reliable curiosity, opening up avenues for curiosity-based exploration in dynamic scenes as generative world models mature.} %

%% file: sec/3_background.tex
\section{Background}
\label{sec:background}

\subsection{Proximal Policy Optimization}
\label{sec:background-ppo}

Proximal Policy Optimization (PPO)~\cite{ppo} is an on-policy actor-critic algorithm that optimizes a policy using trajectories sampled from the current policy. 
Given a policy $\policy_{\policyparams}(\action_t \mid \state_t)$ over actions $\action_t$ conditioned on state $\state_t$, a value function $\valuefn_{\policyparams}(\state_t)$, and advantage estimates $\advantage_t$, PPO updates the policy by maximizing the clipped surrogate objective
\begin{equation}
    \ppoloss(\policyparams)
    =
    \expect_t
    \left[
    \min
    \left(
    \pporatio_t(\policyparams)\advantage_t,
    \mathrm{clip}(\pporatio_t(\policyparams), 1-\clipparam, 1+\clipparam)\advantage_t
    \right)
    \right],
    \qquad
    \pporatio_t(\policyparams)
    =
    \frac{
    \policy_{\policyparams}(\action_t \mid \state_t)
    }{
    \oldpolicy(\action_t \mid \state_t)
    }.
\end{equation}
Here, $\pporatio_t(\policyparams)$ is the policy probability ratio between the updated policy and the rollout policy, and $\clipparam$ is the clipping threshold. 
The clipping term limits the size of policy updates, improving training stability. 
The critic is trained to predict a discounted return target $\ret_t$, formed from accumulated future rewards and value bootstrapping~\cite{gae}. 
Entropy regularization further encourages action diversity, where $\entropy(\policy_{\policyparams}(\cdot \mid \state_t))$ denotes the entropy of the policy distribution at state $\state_t$.
\begin{equation}
    \criticloss(\policyparams)
    =
    \expect_t
    \left[
    \left(
    \valuefn_{\policyparams}(\state_t) - \ret_t
    \right)^2
    \right],
    \qquad
    \entropyloss(\policyparams)
    =
    -
    \expect_t
    \left[
    \entropy
    \left(
    \policy_{\policyparams}(\cdot \mid \state_t)
    \right)
    \right].
\end{equation}

\subsection{3D Gaussian Splatting}
\label{sec:background-3dgs}

3D Gaussian Splatting (3DGS)~\cite{kerbl2023gaussians} represents a scene as a set of anisotropic 3D Gaussian primitives. 
Each Gaussian has a 3D mean $\x_l$, covariance $\covmat_l$, opacity $\opacity_l$, and view-dependent color $\colr_l$. 
During rendering, the Gaussians are projected into the image plane and alpha-composited in depth order to produce the pixel color:
\begin{equation}
    \img(\pixel)
    =
    \sum_{l=1}^{\numgaussians}
    \transmittance_l(\pixel)\,
    \opacity_l\,
    \gaussianproj_l(\pixel)\,
    \colr_l,
    \qquad
    \transmittance_l(\pixel)
    =
    \prod_{j<l}
    \left(
    1-\opacity_j\gaussianproj_j(\pixel)
    \right),
\end{equation}
where $\pixel$ is an image-plane pixel location, $\gaussianproj_l$ is the projected 2D Gaussian footprint, and $\transmittance_l$ is the accumulated transmittance from closer Gaussians. 
The Gaussian parameters are differentiable with respect to an image reconstruction loss, enabling efficient optimization of an explicit 3D scene representation. 
In RGB-D settings, depth can be used to initialize Gaussian means directly by unprojecting observed pixels into 3D, reducing geometric ambiguity and making online optimization faster and more stable, as used in RGB-D Gaussian SLAM systems such as SplaTAM~\cite{splatam}.